\documentclass[runningheads]{llncs}
\usepackage{graphicx}

\usepackage{tikz}
\usepackage{comment}
\usepackage{amsmath,amssymb} 
\usepackage{color}
\usepackage{multirow}
\usepackage{hyperref}

\begin{document}
\pagestyle{headings}
\mainmatter
\def\ECCVSubNumber{5905}  

\title{ClearPose: Large-scale Transparent Object Dataset and Benchmark} 

\titlerunning{ClearPose: Large-scale Transparent Object Dataset and Benchmark}
%
\author{
Xiaotong Chen
\and 
Huijie Zhang
\and
Zeren Yu
\and 
Anthony Opipari
\and \\
Odest Chadwicke Jenkins
}
%
\authorrunning{X. Chen et al.}
%
\institute{University of Michigan, Ann Arbor MI 48109, USA\\
\email{\{cxt,huijiezh,yuzeren,topipari,ocj\}@umich.edu}
}
\maketitle

\begin{abstract}
Transparent objects are ubiquitous in household settings and pose distinct challenges for visual sensing and perception systems.
The optical properties of transparent objects leave conventional 3D sensors alone unreliable for object depth and pose estimation.
These challenges are highlighted by the shortage of large-scale RGB-Depth datasets focusing on transparent objects in real-world settings.
In this work, we contribute a large-scale real-world RGB-Depth transparent object dataset named \textit{\textbf{ClearPose}} to serve as a benchmark dataset for segmentation, scene-level depth completion and object-centric pose estimation tasks.
The ClearPose dataset contains over 350K labeled real-world RGB-Depth frames and 5M instance annotations covering 63 household objects.
The dataset includes object categories commonly used in daily life under various lighting and occluding conditions as well as challenging test scenarios such as cases of occlusion by opaque or translucent objects, non-planar orientations, presence of liquids, etc.
We benchmark several state-of-the-art depth completion and object pose estimation deep neural networks on ClearPose. The dataset and benchmarking source code is available at \href{https://github.com/opipari/ClearPose}{https://github.com/opipari/ClearPose}.

\end{abstract}

\section{Introduction}

Transparent and translucent objects are prevalent in daily life and household settings. When compared with opaque and Lambertian objects, they present additional challenges to visual perception systems. The first challenge is that transparent objects do not exhibit consistent RGB color features across varying scenes. Since the visual appearance of these objects depends on a given scene's background, lighting, and organization, their visual features can drastically differ between scenes thereby confounding feature-based perception systems. The second challenge is the inaccurate depth measurements among RGB-Depth (RGB-D) cameras on transparent or translucent materials due to the lack of reliable reflections. This challenge is especially meaningful for state-of-the-art pose estimation approaches that require accurate depth measurements as input. To overcome these challenges, computational perception algorithms have been proposed for a variety of visual tasks, including image segmentation, depth completion, and object pose estimation. In this paper, our aim is to complement recent work in transparent object perception by providing a large-scale, real-world RGB-D transparent object dataset. Furthermore, we use the new large-scale dataset to perform benchmark analysis of state-of-the-art perception algorithms on transparent object depth completion and pose estimation tasks.

\begin{figure}[h]
    \centering
    \includegraphics[width=\textwidth]{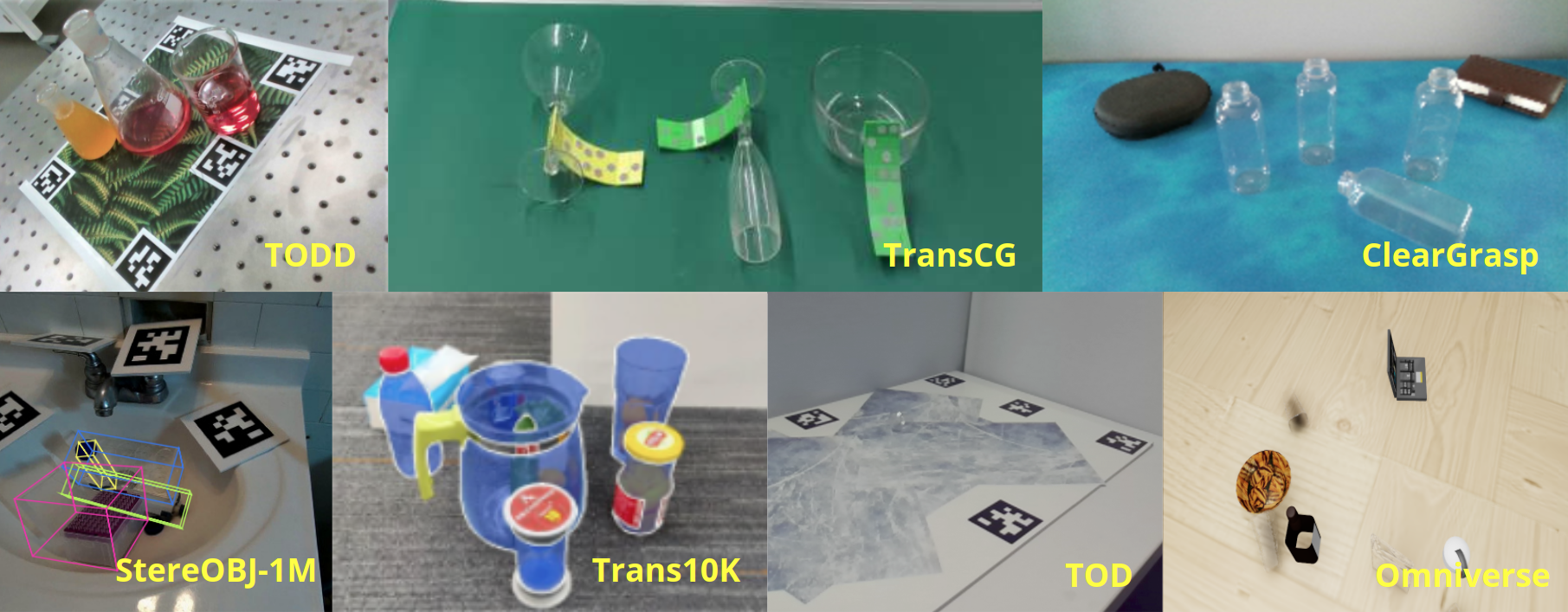}
    \caption{Sample images from existing transparent datasets. TOD and StereOBJ-1M are collected using stereo RGB cameras, and other datasets used RGB-D cameras (except for Omniverse which is completely synthetic).}
    \label{fig:existing_dataset}
\end{figure}

\begin{table}[!h]
\caption{Comparison between existing transparent object datasets and ClearPose. ${^*}$Trans10K is a transparent segmentation dataset and has no object-centric pose labels. ${^*}$StereObj1M is not publicly available at the time of submission so the \#frame and \#pose annotation are estimated based on the published ratio of transparent objects in the entire object set~\cite{liu2021stereobj}}
\resizebox{\textwidth}{!}{%
    \centering
    \begin{tabular}{l|c|c|c|c}
    \hline
    dataset                                 & modality      & \#obj & \#frame               & \#pose annotation\\
    \hline\hline
    TOD~\cite{liu2020keypose}               & RGB-D         & 15    & 48K(real)             & $\sim$0.1M\\
    \hline
    ClearGrasp~\cite{sajjan2020clear}       & RGB-D         & 10    & 50K(syn)+286(real)    & $\sim$0.2M\\  
    \hline
    TODD~\cite{xu2021seeing}                & RGB-D         & 6     & 15K(real)             & $\sim$0.1M\\
    \hline
    Omniverse~\cite{zhu2021rgb}             & RGB-D         & 9     & 60K(syn)              & $\sim$0.2M\\
    \hline
    TransCG~\cite{fang2022transcg}          & RGB-D         & 51    & 58K(real)             & $\sim$0.2M\\  
    \hline
    Trans10K${^*}$~\cite{xie2020segmenting} & RGB           & 10K   & 15K(real)             & seg only\\
    \hline
    ProLIT~\cite{zhou2020lit}               & Light-Field   & 5     & 75K(syn)+300(real)    & $\sim$0.1M\\
    \hline
    StereObj1M${^*}$~\cite{liu2021stereobj} & Stereo        & 7     & $\sim$150K(real)      & $\sim$0.6M\\  
    \hline
    \textbf{ClearPose (ours)}               & RGB-D  & 63    & \textbf{350K(real)}          & \textbf{$\sim$5M}\\
    \hline
    \end{tabular}
}
\label{tab:datasets}
\end{table}

\begin{figure}[!htbp]
    \centering
    \includegraphics[width=\textwidth]{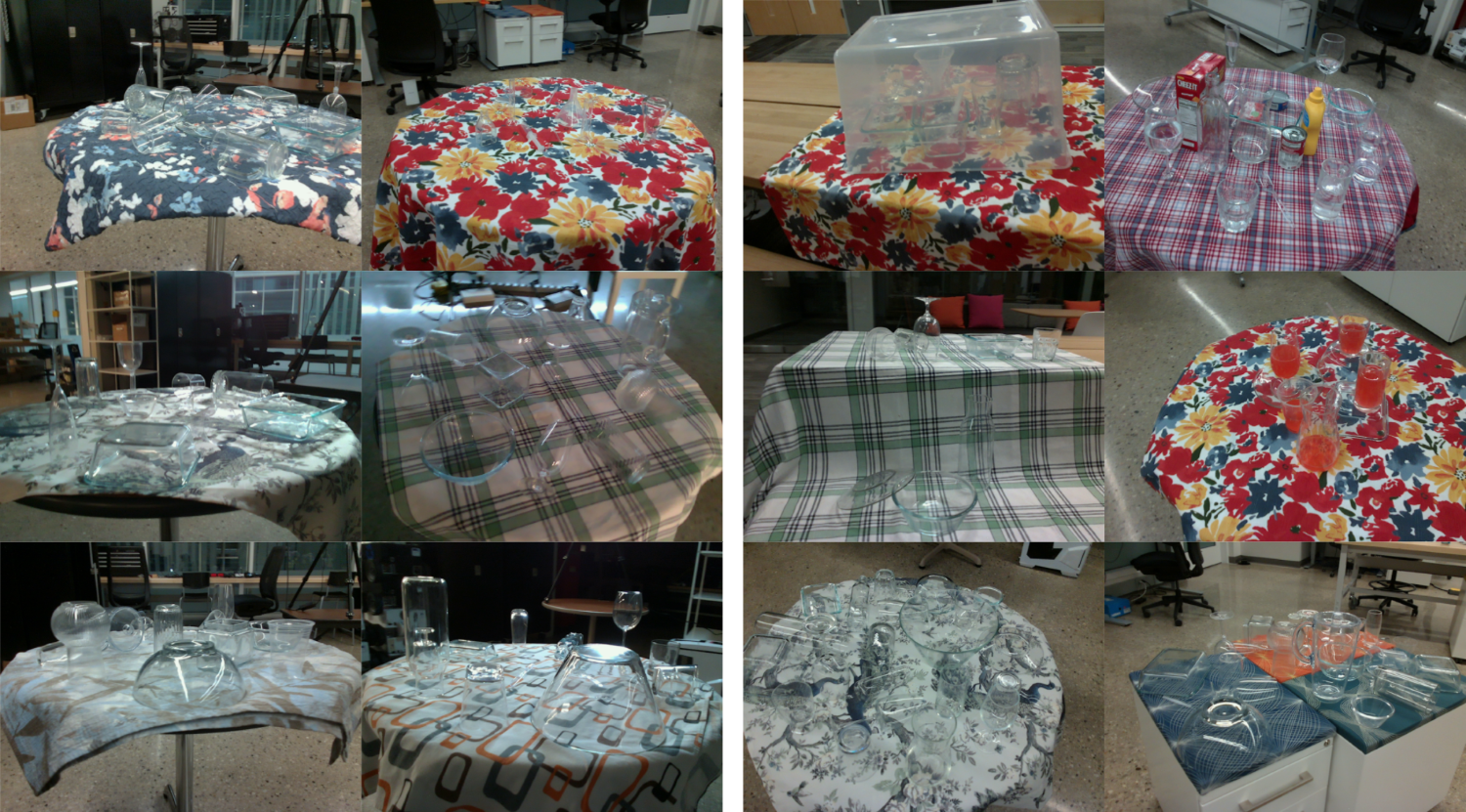}
    \caption{Sample images from ClearPose dataset. On the left, we show RGB images taken for different object subsets under various lighting conditions and backgrounds. On the right, we show examples of different types of testing scenes, such as covered by a translucent box (top-left), novel background with opaque distractor objects (top-right), non-planar cases (middle-left), filled with liquid (middle-right), and heavy clutters of transparent objects (bottom).}
    \label{fig:data_sample}
\end{figure}

There are several existing datasets focusing on transparent object perception with commodity RGB stereo or RGB-D sensors, as summarized in Figure~\ref{fig:existing_dataset}. While these datasets target transparent object perception, most are relatively small-scale (no more than 50K real-world frames), include few cluttered scenes (typically with less than 3 objects per image), do not offer diverse categories of commonplace household objects, and record limited lighting changes. These limitations motivate the introduction of \textbf{ClearPose}, a large-scale real-world transparent object dataset labeled with ground truth pose, depth, instance-level segmentation masks, surface normals, etc. that (1) has a comparable size with ordinary opaque object pose estimation datasets like YCB-Video~\cite{xiang2017posecnn}; (2) contains challenging heavy clutter scenes including multiple layers of occlusion between transparent objects; (3) contains a variety of commonplace household object categories (bottle, cup, wine, container, bowl, plate, spoon, knife, fork, and some chemical lab supplies); (4) covers diverse lighting conditions and multiple adversarial testing scenes. Further details of ClearPose in relation to existing datasets are included in Table.~\ref{tab:datasets} with sample images from ClearPose shown in Figure~\ref{fig:data_sample}.

The labeling of such a large-scale dataset requires both efficiency and accuracy. To achieve these qualities in ClearPose, we take advantage of a recently published pipeline named ProgressLabeller~\cite{progress2022chen}. The ProgressLabeller pipeline combines visual SLAM and an interactive graphical interface with multi-view silhouette matching-based object alignment to enable efficient and accurate labeling of the transparent object poses from RGB-D videos, which realizes rapid data annotation and exempts from the broken depth problem by transparent objects.
Given the unique scale and relevance of ClearPose, we envision it will serve beneficial as a benchmark dataset on transparent perception tasks. In this current paper, we include benchmark analysis for a set of state-of-the-art visual perception algorithms on ClearPose. We target our benchmark analysis on the tasks of depth completion and RGB-D pose estimation.

\section{Related Works}

\subsection{Transparent Dataset and Annotation}

As mentioned in Table~\ref{tab:datasets}, there are several existing datasets and associated annotation pipelines that focus on transparent objects. With the exception of works such as Trans10k~\cite{xie2020segmenting}, TransCut~\cite{xu2015transcut} and TOM-Net~\cite{chen2018tom} that are focused on 2D image segmentation or matting, most recent transparent perception datasets are collected in an object-centric 3D setting using RGB-D, stereo or light-field sensor modalities.

Similar to the case of opaque objects, datasets created in simulation, supporting photo-realistic rendering with ray-tracing, are more readily created and can produce very realistic examples of transparent object appearance. Examples of such simulated datasets include those~\cite{sajjan2020clear} rendered by Blender,~\cite{zhou2020lit} by Unreal Engine and~\cite{zhu2021rgb} by Nvidia Omniverse platform. While simulated datasets are appealing for their ease of creation, they often lack realistic feature artifacts (e.g. sensor noise, object wear marks, true lighting etc.) which can impact downstream perception systems that rely on the synthetic dataset (i.e. the syn-to-real gap).

Among existing real-world datasets, TOD~\cite{liu2020keypose} and StereObj1M~\cite{liu2021stereobj} use RGB stereo cameras together with AprilTags. First, camera pose transforms are solved from AprilTag detection, and then several 2D keypoints on the objects are manually annotated in keyframes that are farthest to each other in the sequence. The corresponding 3D keypoint positions are solved by multi-view triangulation from those labeled 2D keypoints, and finally, the object 6D poses are solved from 3D keypoints as an Orthogonal Procrustes problem and propagated to all frames. In TOD, the authors also introduced a method to record ground truth depth images: they record the positions of transparent objects in the scene, and put their opaque counterparts that share the same shape at the same pose in separate collects. This pipeline was also used for real-world data collection in ClearGrasp~\cite{sajjan2020clear}, however, it's extremely inefficient to replace transparent objects and their counterparts repetitively. Moreover, it is unclear how or whether this approach could be applied to data collection in complex scenes with cluttered objects as is typical in household settings. In Xu et al.~\cite{xu2021seeing}, transparent objects are placed in several fixed locations relative to AprilTag arrays, with pose distribution diversity achieved by attaching a camera to a robot end-effector. This method still restricts the relative position between objects and is inefficient for complex scenes. In Fang et al.~\cite{fang2022transcg}, all objects are attached to a large visual IR marker so that an optical tracking algorithm can estimate the objects' 6D poses. In this way, the collection can support dynamic scenes. On the other hand, all the object instances in collected data are accompanied by visually obscuring external labels which may not exist in natural environments. 

Overall, datasets except for StereObj1M are still not large-scale and require external efforts on hardware, such as deploying robotic arms, calibration between multiple sensors, fiducial or optical markers. Instead of using markers or complex robotic apparatuses, we take advantage of an existing labeling system, ProgressLabeller~\cite{progress2022chen}, that is based on visual SLAM to produce accurate camera poses efficiently on recorded RGB-D videos. There are two assumptions in our labeling pipeline, both of which can be easily met: our pipeline assumes static scenes during video capturing and that scene backgrounds have adequate RGB features for visual SLAM processing.

\subsection{Transparent Depth Completion and Object Pose Estimation}

Zhang et al.~\cite{zhang2018deep} presented early work on the problem of depth completion from inaccurate depth using deep neural networks. Zhang et al. introduced an approach to estimate surface normal and boundaries from RGB images that then solved for the completed depth using optimization. ClearGrasp~\cite{sajjan2020clear} adapted the method to work for transparent objects and demonstrated robotic grasping experiments on transparent objects from completed depth. Tang et al.~\cite{tang2021depthgrasp} integrated the ClearGrasp network structure with adversarial learning to improve depth completion accuracy. Zhu et al.~\cite{zhu2021rgb} proposed a framework that learns local implicit depth functions from the inspiration of neural radiance fields and performs self-refinement to complete the depth of transparent objects. Xu et al.~\cite{xu2021seeing} proposed to first complete the point cloud by projecting the original depth using a 3D encoder-decoder U-Net and then re-project the completed point cloud back to depth, and finally complete this depth using another encoder-decoder network given the ground truth mask. Fang et al.~\cite{fang2022transcg} also used a U-Net architecture to perform depth completion and demonstrated robotic grasping capabilities with their approach.

KeyPose~\cite{liu2020keypose} was proposed for keypoint-based transparent object pose estimation on stereo images. It outperformed DenseFusion~\cite{wang2019densefusion}, even with ground truth depth, and achieved high accuracy on the TOD dataset. Chang et al.~\cite{chang2021ghostpose} proposed a 3D bounding box representation and reported results comparable to KeyPose in multi-view pose estimation. StereObj1M~\cite{liu2021stereobj} benchmarked KeyPose and another RGB-based object pose estimator, PVNet~\cite{peng2019pvnet}, on more challenging objects and scenes, where both methods achieved lower accuracy with respect to the ADD-S AUC metric (introduced in~\cite{xiang2017posecnn}) with both monocular and stereo input. Xu et al.~\cite{xu20206dof} proposed a two-stage pose estimation framework that performs image segmentation, surface normal estimation, and plane approximation in the first stage. The second stage then combines output from the first stage with color and depth RoI features for input to an RGB-D pose estimator originally designed for opaque objects~\cite{tian2020robust} to regress 6D object poses. This method also outperformed DenseFusion and~\cite{tian2020robust} fed with ClearGrasp output depth by a large margin. In this paper, we evaluate how well state-of-the-art RGB-D methods~\cite{he2021ffb6d} designed for opaque object pose estimation can perform on transparent objects compared with~\cite{xu20206dof}.

\section{Dataset}
\subsection{Dataset Objects and Statistics}
\label{sec:dataset}
\begin{figure}[h]
    \centering
    \includegraphics[width=\textwidth]{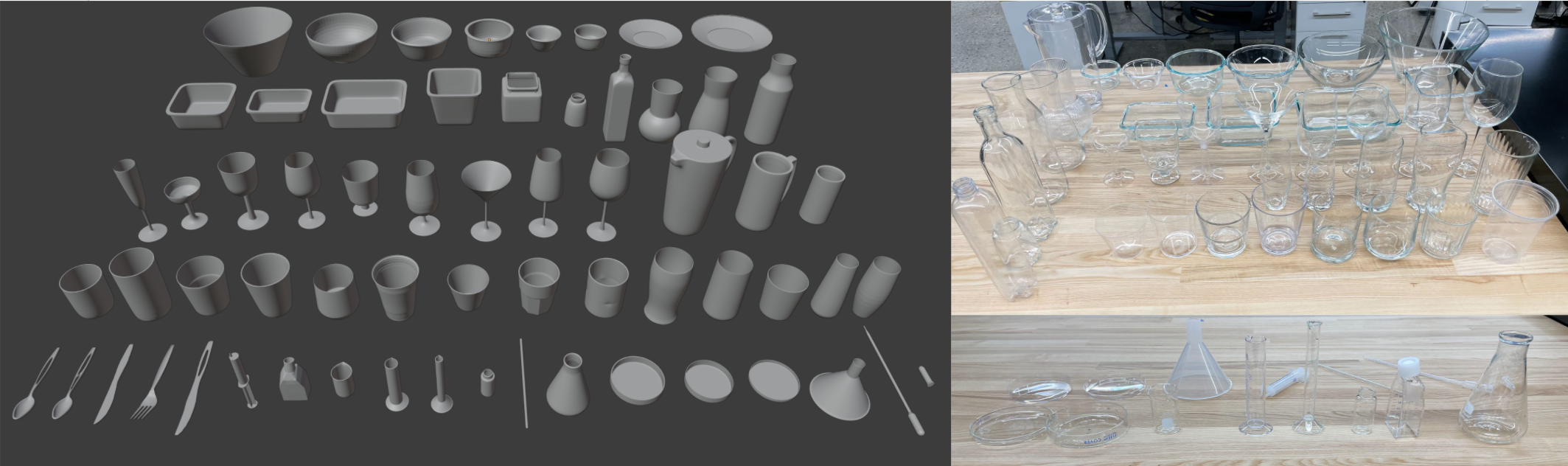}
    \caption{Objects included in the ClearPose dataset. On the left, we show the rendered CAD models for each object. From top to bottom there are bowls and plates (1st row), containers and bottles (2nd row), wine cups, two mugs and a pitcher (3rd row), water cups (4th row), and spoons, a fork, knives as well as chemical supplies (bottom row). On the right, we show real images of household objects on the top right, and chemical supplies on the bottom right.}
    \label{fig:objs}
\end{figure}

As shown in Figure~\ref{fig:objs}, there are 63 objects included in the ClearPose dataset. There are 49 household objects, including 14 water cups, 9 wine cups (with a thin stem compared with water cups), 5 bottles (with an opening smaller than the cross-section of the cylindrical body), 6 bowls, 5 containers (with 4 corners while bowls are classified with round shapes), and several other objects like pitcher, mug, spoon, etc. Moreover, the dataset contains 14 chemical supply objects, including a syringe, a glass stick, 2 reagent bottles, 3 pans, 2 graduated cylinders, a funnel, a flask, a beaker, and 2 droppers.

All the images are collected using a RealSense L515 RGB-depth camera, with a raw resolution of 1280$\times$720. After object pose annotation, the central part of each image is cropped and reshaped to 640$\times$480 for reduced storage space and faster CNN training and inference. For the training set, we separate all 63 objects into 5 separate subsets and collected 4-5 scenes with different backgrounds for each subset. Each scene is scanned by the hand-held camera moving around the tabletop scene at 3 different heights with 3 different lighting conditions (bright room light, dim room light, dim room light with sidelight from a photography lighting board). For the testing set, as the appearance of transparent objects depends on their context within a scene, we consider 6 different test cases and collect corresponding scenes as follows: (1) different backgrounds: novel backgrounds that never appeared in the training scenes with each object subset. (2) heavy occlusions: cluttered scenes each with about 25 objects that form multiple layers of occlusion when viewed from the table's side. (3) translucent/transparent covers: scenes with all transparent objects placed inside a translucent box. (4) together with opaque objects: transparent objects placed together with opaque YCB and HOPE objects, which did not appear in the training set. (5) filled with liquid: scenes with transparent objects filled with different colored liquid. (6) non-planar configuration: scenes with objects placed onto different surfaces with multiple heights. Sample RGB images from both training and test sets are included in Figure~\ref{fig:data_sample}.
\label{sec:data_train_test}

\begin{figure}[t]
    \centering
    \includegraphics[width=\textwidth]{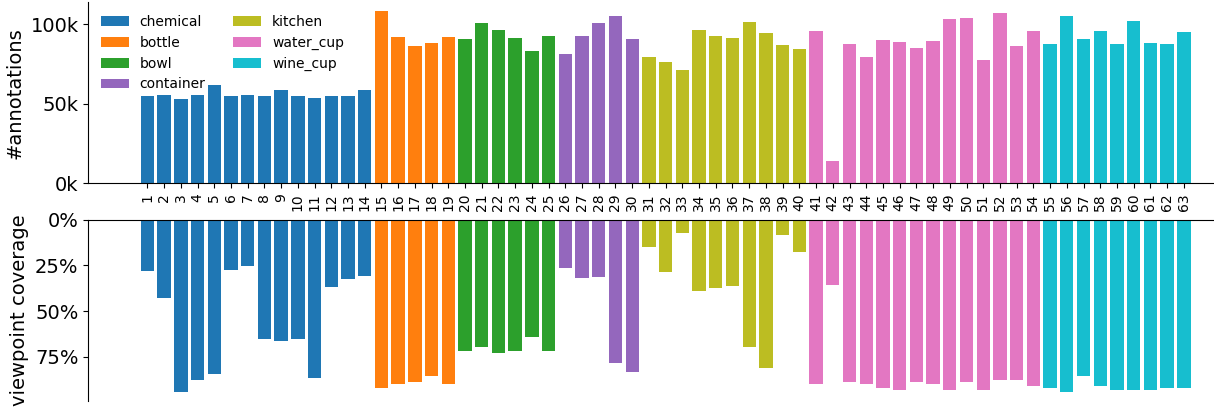}
    \caption{Distribution of instance annotations and viewpoint orientation coverage statistics for every object in the ClearPose dataset. The `kitchen' category includes fork, spoon, knife, mug, plate and pitcher objects. }
    \label{fig:distribution}
\end{figure}

We calculate several statistics about the ClearPose dataset. In total there are 354,481 RGB-D frames captured in 51 scenes, with 5,052,429 object instance annotations with 6 DoF poses, segmentation masks, surface normals, and ground truth depth images. The distribution of object instance annotation and camera viewpoint coverage are shown in Figure~\ref{fig:distribution} colored by object category, where we see our dataset has roughly even viewpoint coverage for most objects. Viewpoint coverage is calculated by projecting collected object orientations onto a unit sphere and counting the covered region percentage over the sphere's surface. For symmetrical objects, regions with the same object appearance are considered together. Some objects like plates, forks, large bowls can only be placed in certain orientations, so they have reduced viewpoint coverage. The 2nd water\_cup was broken during the data collection process so it has fewer instance annotations.

\begin{figure}[t]
    \centering
    \includegraphics[width=\textwidth]{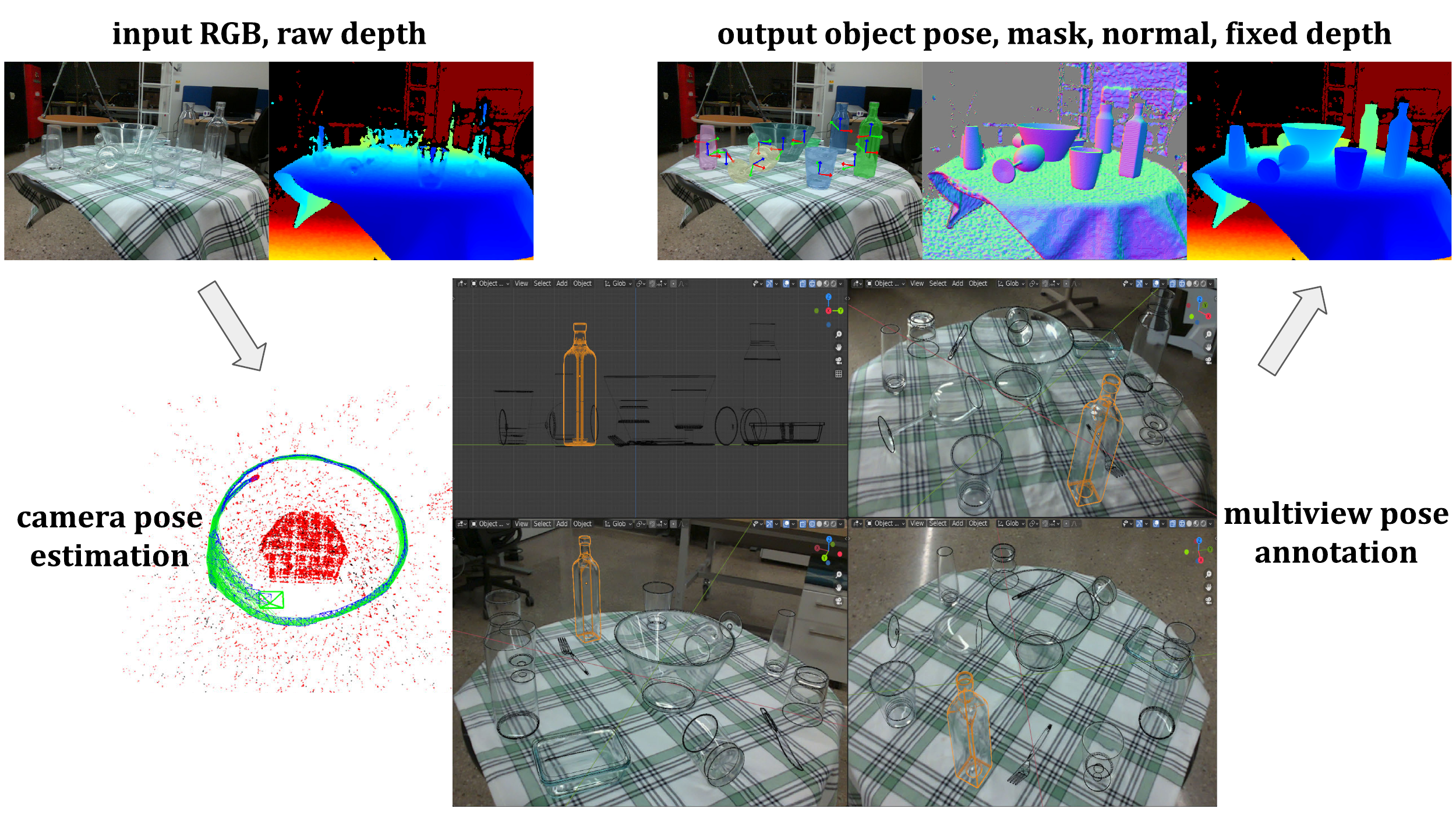}
    \caption{Transparent object pose annotation pipeline using ProgressLabeller. On top-left we show one sample of aligned RGB-D images from continuous streams captured by the camera. The frames are fed into visual SLAM to estimate camera trajectory, then the objects' poses are annotated in a multi-view silhouette alignment interface shown at the bottom-right. Finally, the object poses, surface normals, and fixed depths are calculated and rendered as output.}
    \label{fig:progress_gui}
\end{figure}

\subsection{Pose Annotation}

We use the ProgressLabeller~\cite{progress2022chen} to annotate the 6D poses of transparent objects and render object-wise segmentation masks, ground truth depths, surface normals, etc. from the labeled poses. As shown in Figure~\ref{fig:progress_gui}, the first step of the ProgressLabeller pipeline is to run ORB-SLAM3~\cite{campos2021orb} on collected RGB-D video frames to produce camera pose estimates. During data collection, we notice ORB-SLAM3 sometimes couldn't estimate camera pose well in case of extreme transparent object clutter, where background RGB features are heavily distorted. In these cases, the camera view needs to capture some background area or other landmark objects that can provide stable features. The next step is to import the reconstructed scene (cameras and point cloud) into the Blender workspace, select several camera views from different orientations, and import the object 3D CAD model into the workspace. Then, the object silhouettes/boundaries can be directly compared and matched with original RGB images from multiple views simultaneously when the user drags the object across the scene to tune its position and orientation.  Figure~\ref{fig:progress_gui} shows an example case of a matched transparent bottle. Optionally, the user can select to first locate the object onto the 2D fitting plane of the support table, etc., as shown on top-left of Figure~\ref{fig:progress_gui}. After labeling all objects' poses in the 3D workspace, their poses in every camera frame are calculated by dividing them with estimated camera poses. The output ground truth depth images (fixed depth) are generated by overlaying rendered object CAD model depth to the original depth images. Then the surface normals are calculated from the depth images. It takes around 30 minutes to finish labeling one scene using this pipeline, including visual SLAM for camera pose estimation, object pose manual alignment, and output image rendering.

\section{Benchmark Experiments}
In this section, we provide benchmark results of recent research on depth-related transparent object perception using deep neural networks, including 
scene level depth completion, and both instance-level and category-level RGB-D pose estimation. As mentioned in Section~\ref{sec:data_train_test}, we test the generalization capability of such systems on 6 aspects of appearance novelty with transparent objects: new background, heavy occlusion, translucent cover, opaque distractor objects, filled with liquid, and non-planar placement. Specifically, around 200K images are selected for training, and for each of 6 test cases, 2K images from corresponding scenes are randomly sampled to compose the testing set for evaluation.



\subsection{Depth Completion}

\begin{figure}
    \centering
    \includegraphics[width=\textwidth]{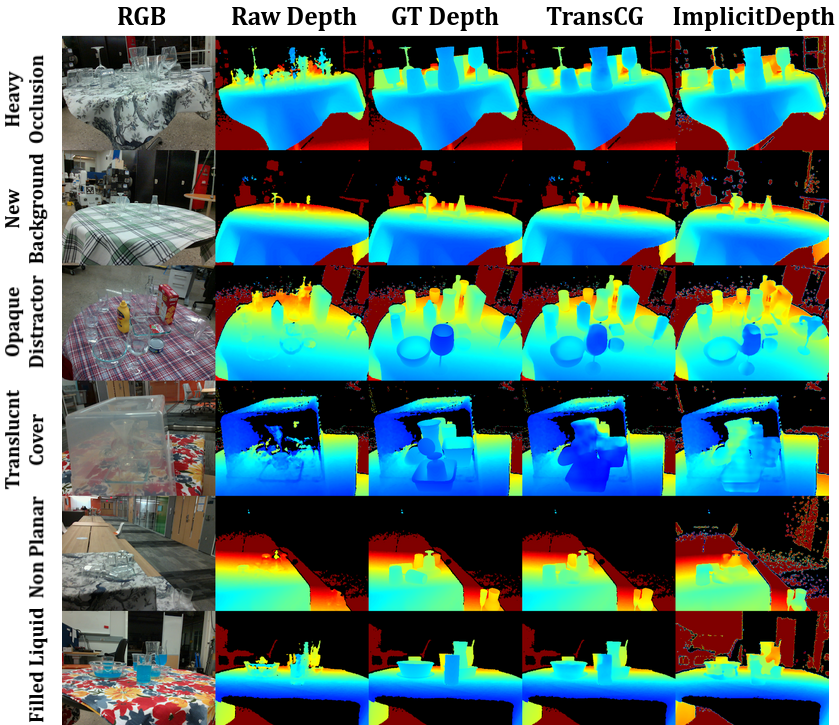}
    \caption{Qualitative depth completion results. From left to right, there are RGB image, raw depth, ground truth depth rendered using object CAD models, completed depth by TransCG and ImplicitDepth.}
    \label{fig:depth_completion}
\end{figure}

We selected two recent depth completion works that are publicly available, ImplicitDepth~\cite{zhu2021rgb} and TransCG~\cite{fang2022transcg} as baseline methods for the depth completion task on transparent object scenes. (ClearGrasp~\cite{sajjan2020clear} was shown to be less accurate than both works, and TranspareNet~\cite{xu2021seeing} was released around the date of this submission.) ImplicitDepth is a two-stage method that learns local implicit depth functions in the first stage through ray-voxel pairs similar to neural rendering and refines the depth in the second stage. TransCG is built on DFNet~\cite{hong2019deep} which was initially developed for image completion. We trained both networks following their original papers' training iterations and hyper-parameters. Specifically, ImplicitDepth was trained on a 16G RTX3080 GPU with a 0.001 learning rate and iterated around 2M frames for each of the two stages. TransCG was trained on an 8G RTX3070 GPU with a fixed 0.001 learning rate and iterated around 900K frames in total. Both works use Adam as the optimizer. Then we evaluated the two works on 6 test sets mentioned in Section~\ref{sec:data_train_test} with metrics defined in~\cite{eigen2014depth}. The results are shown in Table~\ref{tab:depth_completion}. TransCG surpassed ImplicitDepth in most tests with fewer training iterations, which implies that methods using DFNet can outperform designs using voxel-based PointNet for transparent depth completion. Across different tests, both methods perform poorly in Translucent Cover scenes and achieved the best performance in New Background scenes. Other scene variations such as Filled Liquid, Opaque Distractor, and Non Planar do not substantially impact the methods' accuracy. Figure~\ref{fig:depth_completion} shows examples of qualitative results from both methods compared with the ground truth.

\begin{table}[]
\caption{Depth completion benchmark results of ImplicitDepth and TransCG on 6 different test scenarios of the ClearPose dataset.}
\resizebox{\textwidth}{!}{%
\begin{tabular}{l|l|cccccc}
\hline
Testset                            & Metric      & RMSE$\downarrow$ & REL$\downarrow$ & MAE$\downarrow$ & $\delta_{1.05}\uparrow$ & $\delta_{1.10}\uparrow$ & $\delta_{1.25}\uparrow$ \\ \hline
\multirow{2}{*}{New Background}    & ImplicitDepth &  0.07   & 0.05    & 0.04    &  67.00    & 87.03     & 97.50     \\
                                   & TransCG       &  0.03    &  0.03   &  0.02   &  86.50    &   97.02 &  99.74    \\ \hline
\multirow{2}{*}{Heavy Occlusion}   & ImplicitDepth & 0.11  & 0.09  & 0.08  & 41.43 & 66.52 & 91.96     \\
                                   & TransCG       &   0.06   &   0.04  & 0.04    &  72.03    & 90.61     &  98.73    \\ \hline
\multirow{2}{*}{Translucent Cover} & ImplicitDepth &  0.16 &  0.16 &  0.13  &22.85  &41.17  &73.11  \\
                                   & TransCG       & 0.16   & 0.15 &0.14 &  23.44   & 39.75     &67.56 \\ \hline
\multirow{2}{*}{Opaque Distractor} & ImplicitDepth & 0.14 &0.13   & 0.10    &34.41  &55.59 & 83.23 \\
                                   & TransCG       & 0.08  &  0.06 &0.06 &52.43 &  75.52   &  97.53\\ \hline
\multirow{2}{*}{Filled Liquid}     & ImplicitDepth &0.14  &0.13 & 0.11&  32.84&  53.44&  84.84\\
                                   & TransCG       &0.04 &0.04 & 0.03    & 77.65     &    93.81  & 99.50     \\ \hline
\multirow{2}{*}{Non Planar}        & ImplicitDepth &   0.18   &    0.16 &     0.15& 20.34     & 38.57     & 74.02     \\
                                   & TransCG       &  0.09    & 0.07    & 0.07    & 55.31     & 76.47     & 94.88     \\ \hline
\end{tabular}%
}
\label{tab:depth_completion}
\end{table}

\subsection{Instance-Level Object Pose Estimation}
There is one recent work of Xu et al.~\cite{xu20206dof} focusing on transparent object pose estimation using raw RGB and depth images. This work doesn't have source code publicly available, so we re-implemented the proposed method following the original paper for inclusion in our benchmark analysis. This method is implemented as a two-stage pipeline, for which we trained Mask R-CNN~\cite{he2017mask} for instance-level segmentation and DeepLabv3~\cite{chen2017rethinking} for surface normal estimation, and with an XYZ 3D coordinate map of the supporting plane feature in the first stage. The second stage replicates most of the architecture and loss functions described in~\cite{tian2020robust} to ultimately regress dense pixel-wise 3D translation, 3D rotation delta from a set of fixed rotation anchors, and confidence scores. In practice, we trained the networks on an RTX2080-SUPER GPU. Mask R-CNN has trained 5 epochs with SGD optimizer, batch size of 5, and learning rate of 0.005. DeepLabv3 was trained 2 epochs with Adam optimizer, batch size of 4 and learning rate of 0.0001, and second stage network was trained around 180K iterations with Adam optimizer, batch size of 4, and learning rate of 0.0005. We compare this method~\cite{xu20206dof} with a state-of-the-art RGB-D pose estimator that was originally designed for opaque objects, FFB6D~\cite{he2021ffb6d}. The FFB6D estimator follows a two-stream RGB and point cloud encoder-decoder architecture with fusion between blocks. FFB6D is trained on a 16G RTX3080 GPU for 5 epochs with batch size 6. All the hyper-parameters follow the default value from the original implementation.  For our analysis of FFB6D pose estimation performance, we run experiments with different depth options in training and testing: with raw, ground truth, and completed depth from TransCG, as detailed in Table~\ref{tab:inst_pose_estimation}. 

\begin{table}[h]
\caption{Pose estimation accuracy comparison on different test sets of the ClearPose dataset. $\textrm{FFB6D}_{r/r}$ refers to train and test FFB6D model both on raw depth. Similarly, $\textrm{FFB6D}_{g/c}$, $\textrm{FFB6D}_{g/g}$ refer to train on ground truth, test on completed depth from TransCG, and train and test both on ground truth, respectively. The values are averaged across all objects in the dataset.}
\resizebox{\textwidth}{!}{%
\begin{tabular}{l|l|cccc}
\hline
Testset & Metric & Xu et al. & $\textrm{FFB6D}_{r/r}$ &  $\textrm{FFB6D}_{g/c}$ &  $\textrm{FFB6D}_{g/g}$ \\ \hline
\multirow{2}{*}{New Background}    & Accuracy   &  50.958  &   44.264     &   49.517     &  59.694 \\ \cline{2-6} 
                                   & ADD(-S)    &  45.233  &   43.452     &   47.691     & 58.224 \\ \hline 
\multirow{2}{*}{Heavy Occlusion}   & Accuracy   &  24.193  &   14.723     &   15.160     &  26.331 \\ \cline{2-6} 
                                   & ADD(-S)    &  22.953  &   17.869     &   17.862     & 31.875  \\ \hline
\multirow{2}{*}{Translucent Cover} & Accuracy   &  14.353  &   5.5617      & 4.5345        & 13.433  \\ \cline{2-6} 
                                   & ADD(-S)    &  14.311  &   7.5983      & 5.8054        & 17.620  \\ \hline
\multirow{2}{*}{Opaque Distractor} & Accuracy   &  42.630  &   0.4618      & 1.3331        & 2.3525  \\ \cline{2-6} 
                                   & ADD(-S)    &  39.036  &   0.7628      & 1.5516        & 3.0685  \\ \hline
\multirow{2}{*}{Filled Liquid}     & Accuracy   &  34.500  &   7.6908      & 9.0584        & 16.228  \\ \cline{2-6} 
                                   & ADD(-S)    &  32.251  &   11.153     & 10.828       & 18.583  \\ \hline
\multirow{2}{*}{Non Planar}        & Accuracy   &  21.024  &   7.4924      & 7.5843        & 15.567 \\ \cline{2-6} 
                                   & ADD(-S)    &  18.411  &   7.8021      & 6.7339        & 16.986  \\ \hline
\end{tabular}%
}
\label{tab:inst_pose_estimation}
\end{table}

For evaluation, we use two metrics based on Average-point-Distance (ADD and ADD-S) from~\cite{xiang2017posecnn}. ADD is calculated as the average Euclidean distance of corresponding point pairs from two object point clouds separately at the ground truth pose and predicted pose. ADD-S is calculated as the minimum distance of every point from the predicted point cloud to the ground truth point cloud. In Table~\ref{tab:inst_pose_estimation}, `Accuracy' is calculated as the percentage of all pose estimates on the test set with ADD error less than 10cm. `ADD(-S)' is calculated as Accuracy-Under-Curve area integrated from 0-10cm error, which is then scaled from 1 to 100 as the percentage.

\begin{figure}[!b]
    \centering
    \includegraphics[width=\textwidth]{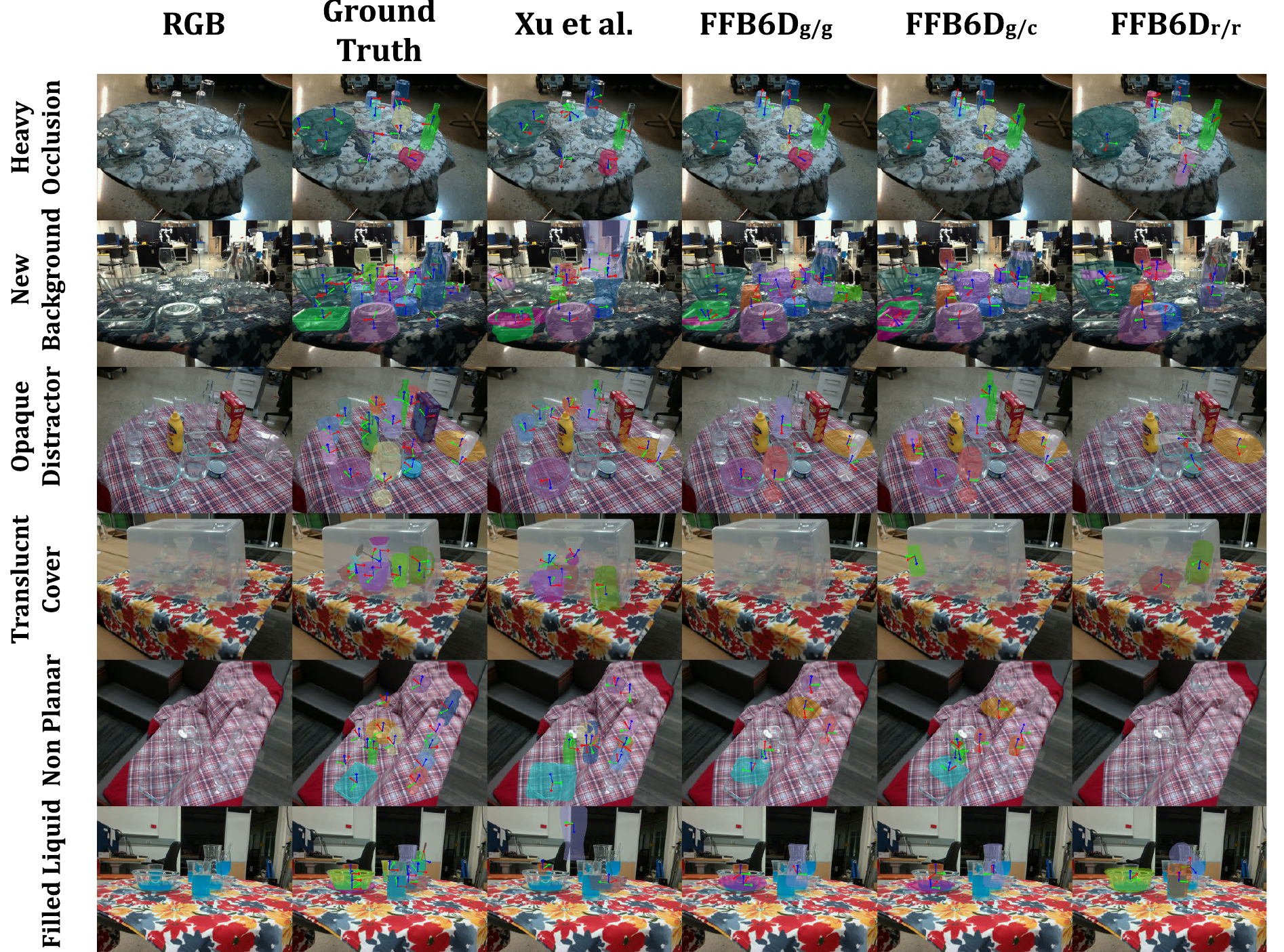}
    \caption{Visualization of pose estimation in ClearPose dataset. From left to right, they are raw image, ground truth object poses, pose estimation results of method from Xu et al., FFB6D$_{g/g}$, FFB6D$_{g/c}$, FFB6D$_{r/r}$. From top to bottom, results are shown in different test scenes in Table~\ref{tab:inst_pose_estimation}. Objects are projected to color masks based on their pose estimates, with their 6DoF poses marked as the red-green-blue coordinate frame.}
    \label{fig:pose_estimation}
\end{figure}

As shown in Table~\ref{tab:inst_pose_estimation}, from the comparison between different training and testing combination within FFB6D, the upper bound performance appears when the network are both trained and tested on ground truth depth. When both the training and testing data come to raw, the metric drops a lot. Obviously, inaccurate depth would be the difficulty for transparent object pose estimation. It should be mentioned that training on ground truth depth, testing on completed depth (from TransCG) almost display the same accuracy. Although TransCG is good at depth completion, the disparity between ground truth and depth completion would make the network in vain. Generally, the easiest test case is New Background, and the accuracy drops a lot in the other 5 scenarios.  When we compare the accuracy of Xu et al. with variants of FFB6D, we find they are comparable in New Background, Heavy Occlusion, Translucent Cover, and Non Planar scenes, while Xu et al. is much better in Opaque Distractor and Filled Liquid scenes. One possible reason is that there are some unseen colors mixing in the transparent objects, adding remarkable noise to object keypoint regression during the FFB6D inference process, which is not used by Xu et al. Overall, the pose estimation accuracy of current methods is still much worse than that on opaque objects with RGB features (with ADD-S around 90 on public datasets~\cite{hodavn2020bop}). Some qualitative examples of pose estimates are shown in Figure~\ref{fig:pose_estimation}.

\section{Discussions}
There are some common classes of objects with transparency/translucency not included in our dataset, for example, those with colored transparent/translucent materials, with markers or labels, together with opaque parts, etc. Instead, our focus in the ClearPose dataset is to investigate pure transparency that exhibits relatively few features for perception.
On the other hand, we anticipate the open-source ProgressLabeller~\cite{progress2022chen} will facilitate more large-scale customized transparent datasets in the future.

\begin{figure}[b]
    \centering
    \includegraphics[width=0.8\textwidth]{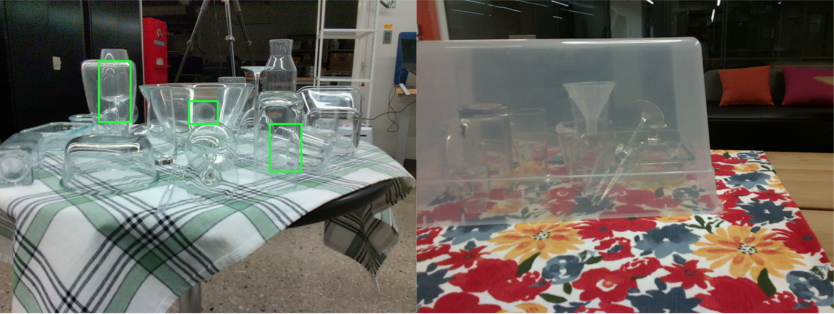}
    \caption{Examples of multi-layer transparent object appearance. In the left image, the annotated bounding boxes show large overlap between object pairs, where the objects behind are still perceivable in some cases with less light distortion. In the right image, objects behind the translucent surface are still detectable as well.}
    \label{fig:multilayer}
\end{figure}

As for benchmarking perception models, we didn't include a complete list of recent state-of-the-art approaches due to resource constraints (i.e. compute and time limitations). Based on our current dataset and benchmark results, there are several possible extensions: (1) Comparison of RGB-only pose estimators with RGB-D methods that are free of transparent object broken depth issues. (2) Category-level pose estimation for transparent objects, for which the ClearPose dataset has categories of bowls, bottles, wine\_cups, etc. that are with similar 3D shape and topology. (3) Neural rendering on transparent objects considering environment contexts, such as varied lighting and occlusions. (4) Transparent object grasping and manipulation experiment in practical scenes, including the 6 test scenarios mentioned in the benchmark.

Besides, an especially interesting problem emerging from our heavy cluttered, and translucent covered test scenes is the multi-layer appearance of transparent objects. As shown in Figure~\ref{fig:multilayer}, because of transparency/translucency, some image pixels could belong to more than one object. New detection and segmentation annotation rules, such as bounding box non-maximum suppression threshold, or segmentation mask format over the image, could be proposed and explored based on our dataset as future work.

\section{Conclusions}
In this paper, we described the contribution of \textbf{ClearPose}, a new large-scale RGB-D transparent object dataset with annotated poses, masks, and associated labels created using a recently proposed pipeline. We performed a set of benchmarking experiments on depth completion and object pose estimation tasks using state-of-the-art methods over 6 different generalization test cases that are common in practical scenarios. Results from our experiments demonstrate that there is still much room for improvement in some cases, such as heavy clutter, transparent objects filled with liquid, or being covered by other translucent surfaces. The dataset and benchmark code implementations will be made public with the intention to support further research progress in transparent RGB-D visual perception.

\noindent
\textbf{Acknowledgement}. We thank greatly the support from Dr. Peter Gaskell and Weishu Wu at the University of Michigan, who provided devices and objects for dataset collection.

\bibliographystyle{splncs04}
\bibliography{egbib}

\title{Supplementary Material}
\titlerunning{Supplementary Material}
%
\author{
Xiaotong Chen
\and 
Huijie Zhang
\and
Zeren Yu
\and 
Anthony Opipari
\and \\
Odest Chadwicke Jenkins
}
%
\authorrunning{X. Chen et al.}
%
\institute{University of Michigan, Ann Arbor MI 48109, USA\\
\email{\{cxt,huijiezh,yuzeren,topipari,ocj\}@umich.edu}
}
\maketitle

\appendix

\section{Dataset Collection Details}
The 3D CAD models of objects in the ClearPose dataset are manually created in Blender, with geometry size measurement of real objects. Then during the labeling process of aligning object models to RGB images, the object dimensions are further verified and corrected. Finally, the object models could align well with RGB images in multiple scene sequences.

\begin{figure}
    \centering
    \includegraphics[width=\textwidth]{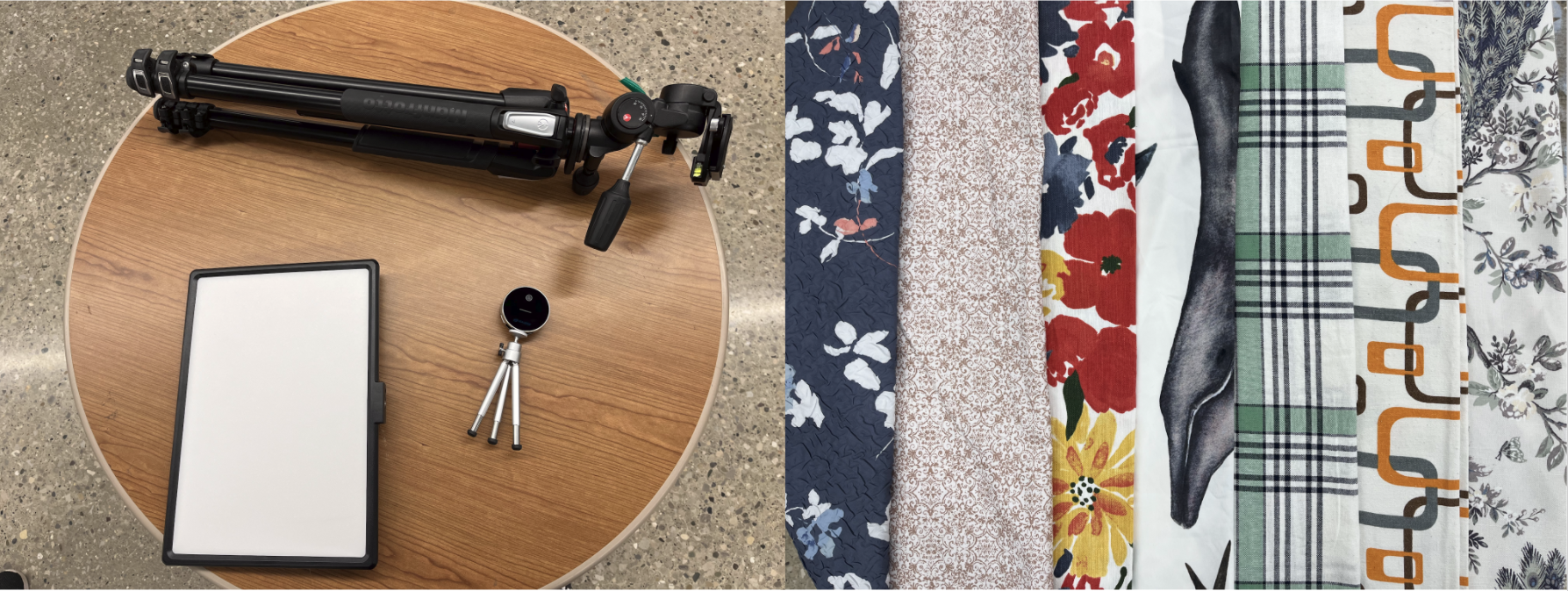}
    \caption{On the left there are lighting board, tripod, and Intel RealSense L515 camera put on the background round table. On the right there are most of background table cloth used in the dataset.}
    \label{fig:bg}
\end{figure}

We used a series of table cloth with different textures as background when collecting the dataset. For each scene, after we randomly put objects in place, a collector will hold the camera and move steadily around the scene with normal room light from relative low, middle, high altitude view angles, for 1 round for each view angle, then another 3 rounds with dim room light and 3 rounds with dim room light plus side light from the lighting board mounted on a tripod on the side. The RealSense camera records aligned RGB-depth image pairs around 30Hz with resolution 1280$\times$720 in scenes with normal room light, and 10-20Hz in other scenes. Most device mentioned above are shown in Figure~\ref{fig:bg}. 

A video with detailed annotation steps and results of depth completion and object pose estimation is available at \href{https://www.youtube.com/watch?v=i8LjxicAaps}{YouTube}

\section{Object Pose Estimation}
More qualitative results with visualization on depth completion and object pose estimation are included in the accompanied video.

\begin{figure}
    \centering
    \includegraphics[width=\textwidth]{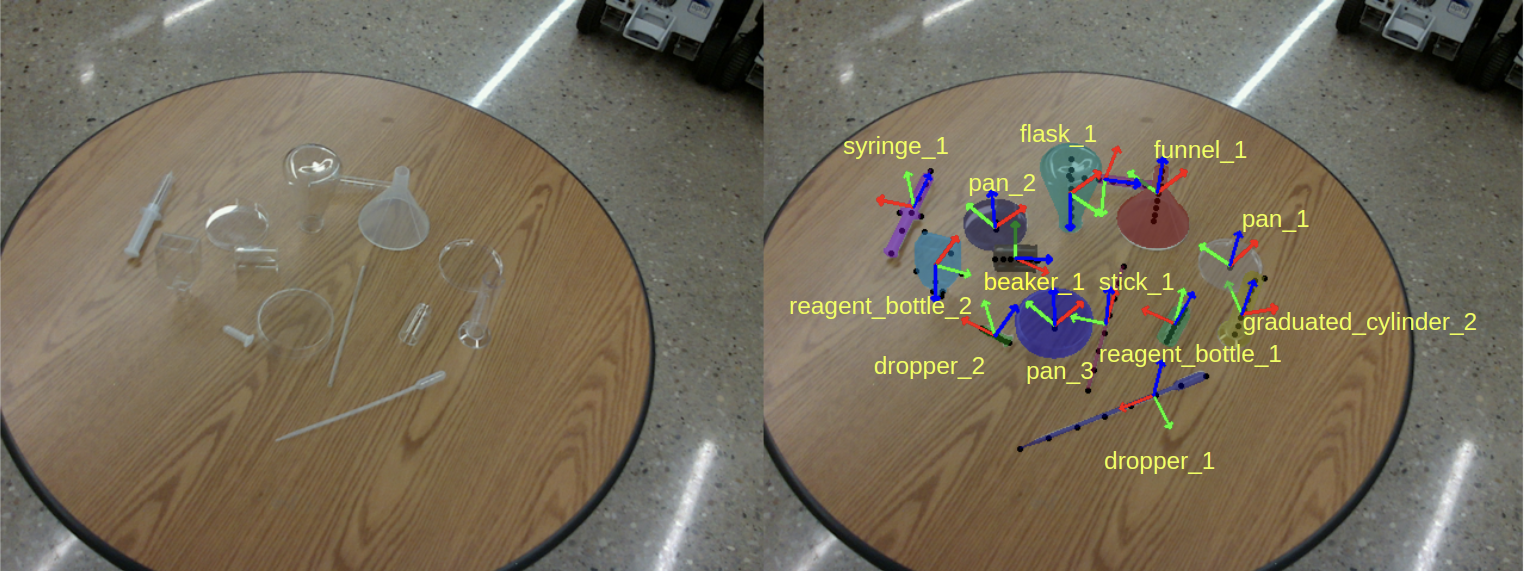}
    \caption{Object names and labeled keypoints in set 1 of 5.}
    \label{fig:kp1}
\end{figure}

\begin{figure}
    \centering
    \includegraphics[width=\textwidth]{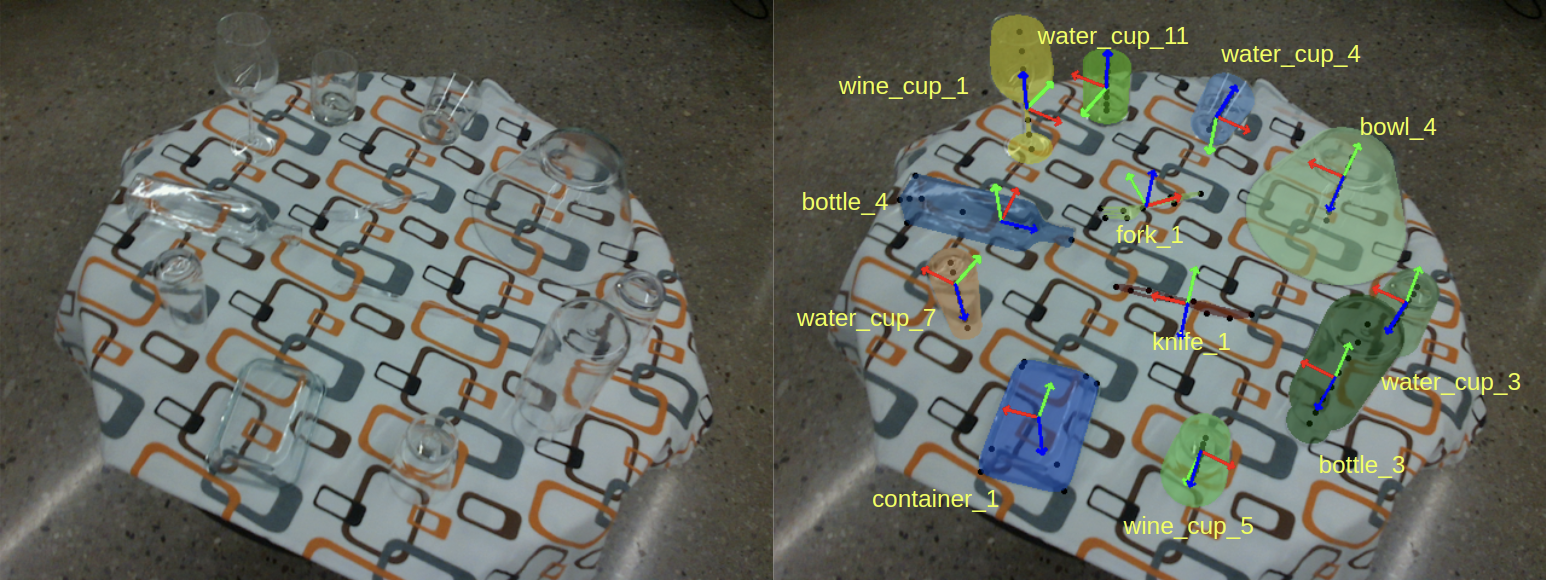}
    \caption{Object names and labeled keypoints in set 2 of 5.}
    \label{fig:kp2}
\end{figure}

\begin{figure}
    \centering
    \includegraphics[width=\textwidth]{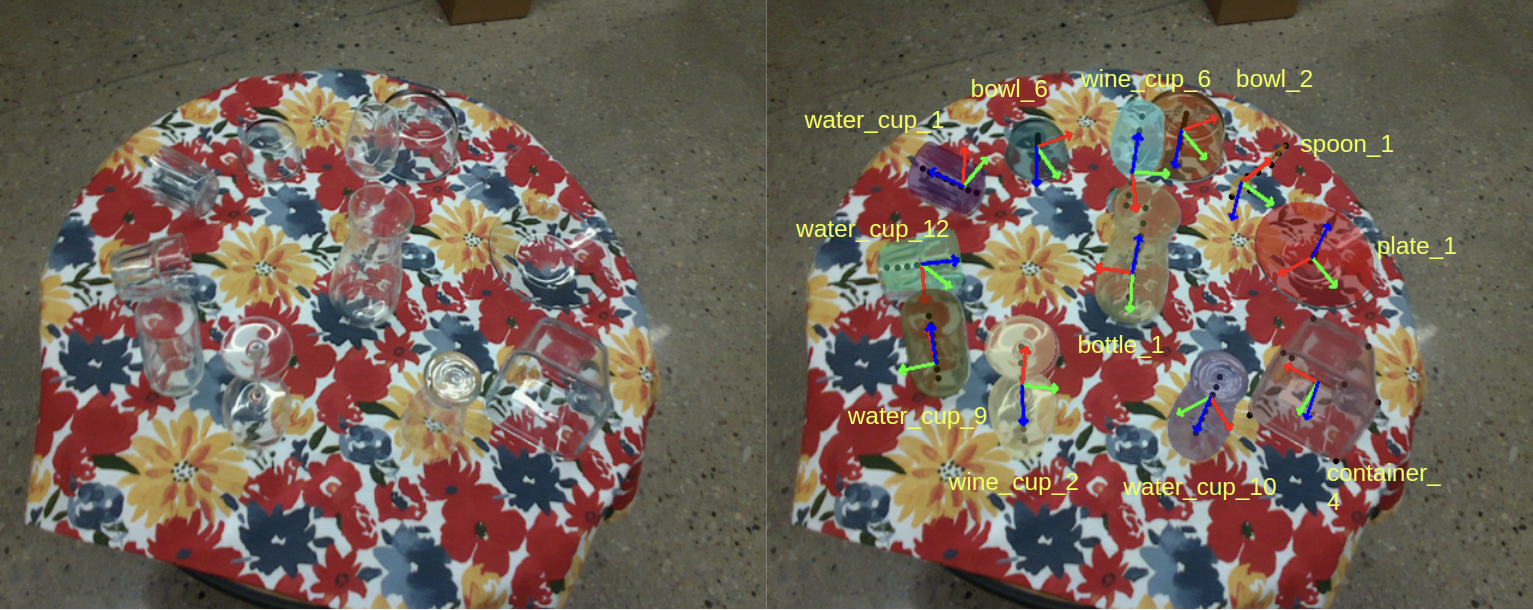}
    \caption{Object names and labeled keypoints in set 3 of 5.}
    \label{fig:kp3}
\end{figure}

\begin{figure}
    \centering
    \includegraphics[width=\textwidth]{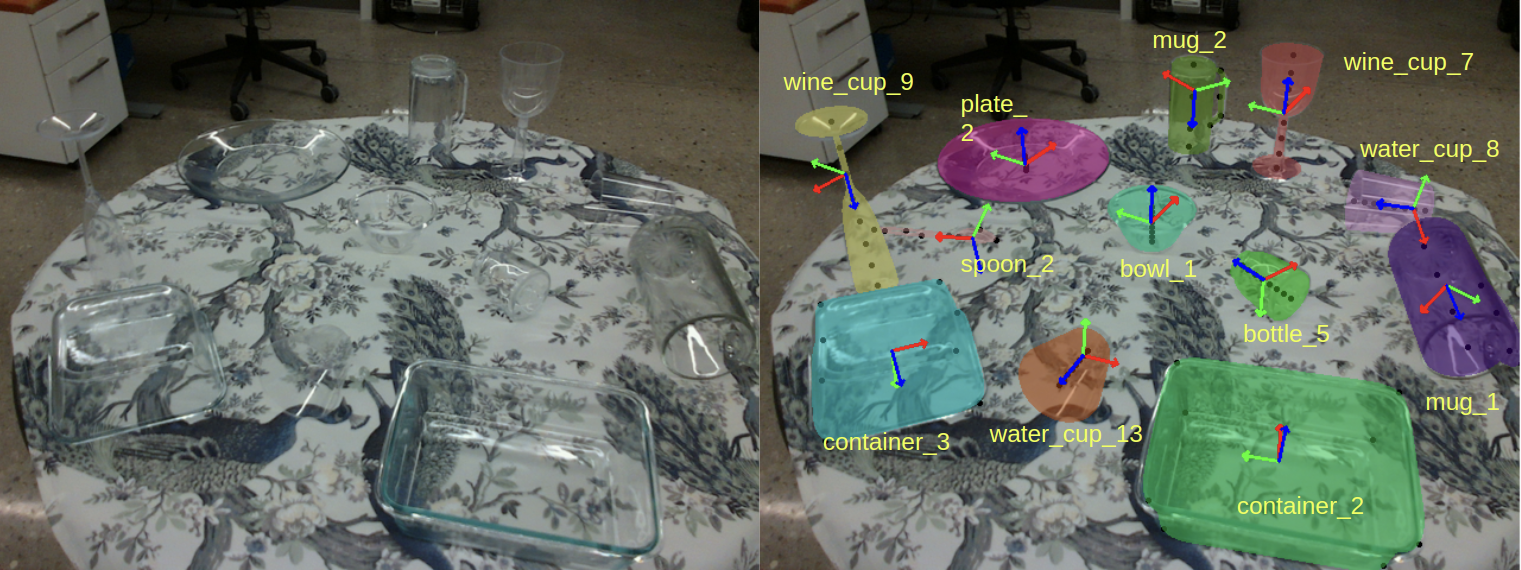}
    \caption{Object names and labeled keypoints in set 4 of 5.}
    \label{fig:kp4}
\end{figure}

\begin{figure}
    \centering
    \includegraphics[width=\textwidth]{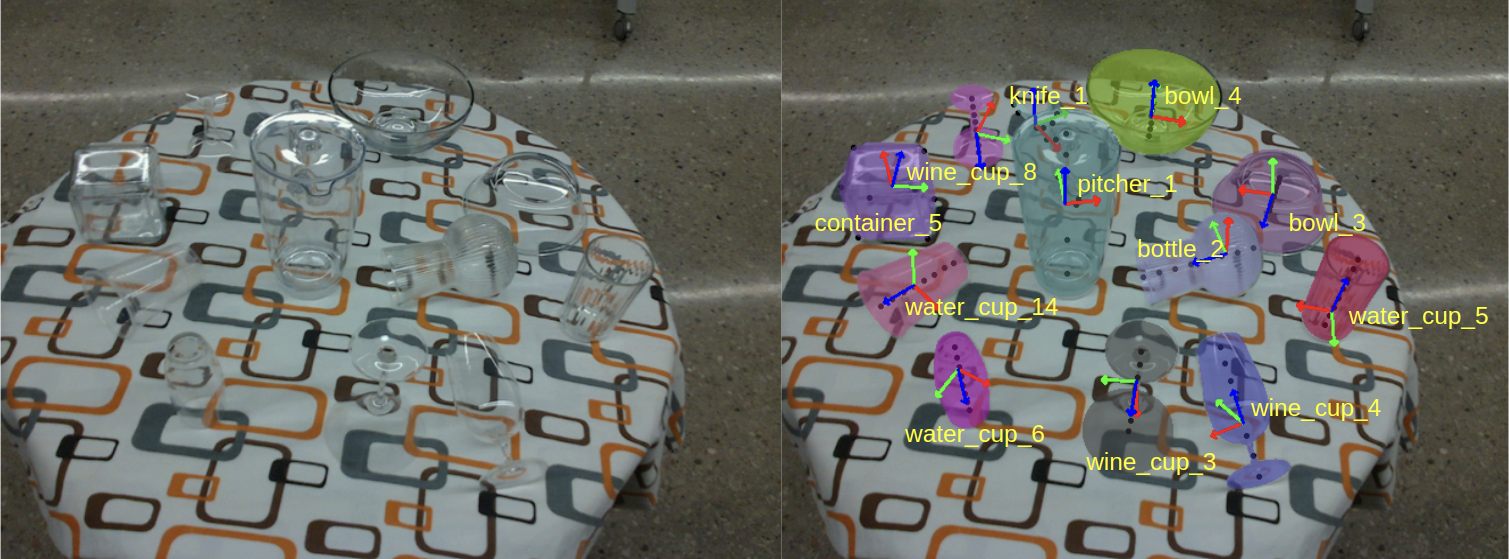}
    \caption{Object names and labeled keypoints in set 5 of 5.}
    \label{fig:kp5}
\end{figure}

Specific to FFB6D, eight 3D sample keypoints on each object are required for the deep neural network to regress. We manually specify keypoints as shown below Figure~\ref{fig:kp1}-\ref{fig:kp5}. For axial-symmetric objects, all of eight keypoints are generated with equal interval along the symmetric axis. For other objects, keypoints are manually selected to be at feature-rich areas, such as corners or edges.

\end{document}